\documentclass[10pt,twocolumn,letterpaper]{article}

\usepackage{iccv}
\usepackage{alex}
\usepackage{times}
\usepackage{epsfig}
\usepackage{graphicx}
\usepackage{amsmath}
\usepackage{amssymb}
\usepackage[ruled]{algorithm2e}
\usepackage[table]{xcolor}
\usepackage{array}
\usepackage{pgfplots}
\usepackage{optidef}
\usepackage{subcaption}
\usepackage[sort]{cite}
\usepackage{tango}
\usepackage[titletoc,title]{appendix}

\pgfplotsset{compat = 1.3}
\usepgfplotslibrary{colorbrewer}
\definecolor{Dark2-3-1}{RGB}{27,158,119}
\definecolor{Dark2-3-3}{RGB}{117,112,179}
\definecolor{Dark2-8-1}{RGB}{27,158,119}
\definecolor{Dark2-8-A}{RGB}{27,158,119}
\definecolor{Dark2-8-2}{RGB}{217,95,2}
\definecolor{Dark2-8-B}{RGB}{217,95,2}
\definecolor{Dark2-8-3}{RGB}{117,112,179}
\definecolor{Dark2-8-C}{RGB}{117,112,179}
\definecolor{Dark2-8-4}{RGB}{231,41,138}
\definecolor{Dark2-8-D}{RGB}{231,41,138}
\definecolor{Dark2-8-5}{RGB}{102,166,30}
\definecolor{Dark2-8-E}{RGB}{102,166,30}
\definecolor{Dark2-8-6}{RGB}{230,171,2}
\definecolor{Dark2-8-F}{RGB}{230,171,2}
\definecolor{Dark2-8-7}{RGB}{166,118,29}
\definecolor{Dark2-8-G}{RGB}{166,118,29}
\definecolor{Dark2-8-8}{RGB}{102,102,102}
\definecolor{Dark2-8-H}{RGB}{102,102,102}
\pgfplotsset{legend cell align=left}
\tikzset{
    posstyle/.style = {draw=skyblue2, line width=3, solid},
    negstyle/.style = {draw=chameleon3, line width=3, densely dashed},
}
\pgfplotscreateplotcyclelist{mycolorlist}{%
skyblue2\\
chameleon3\\
orange1\\
aluminium6\\
scarletred3\\
plum1\\
skyblue3\\
chocolate2\\
maroon\\
butter3\\
}
\pgfplotscreateplotcyclelist{mystylelist}{%
solid\\
densely dashdotted\\
dashed\\
densely dashed\\
dashdotted\\
densely dotted\\
}

\newcommand{\loss}[1]{\ell^{\mathrm{#1}}}
\newcommand{\triplet}{\ell^{\mathrm{triplet}}}
\newcommand{\contrast}{\ell^{\mathrm{contrast}}}

\newcommand{\bglob}{\beta^{(\mathrm{0})}}
\newcommand{\bclass}{\beta^{(\mathrm{class})}}

\newcommand{\bex}{\beta^{(\mathrm{img})}}

\newcommand{\nubglob}{\beta^{(\mathrm{0})}}
\newcommand{\nubclass}{\beta^{(\mathrm{class})}}

\usepackage[pagebackref=true,breaklinks=true,colorlinks,bookmarks=false]{hyperref}

\iccvfinalcopy 

\def\iccvPaperID{1246} 

\ificcvfinal\fi
\begin{document}

\title{Sampling Matters in Deep Embedding Learning}

\author{Chao-Yuan Wu\thanks{Part of this work performed while interning at Amazon.}\\
UT Austin \\
{\tt\small cywu@cs.utexas.edu}
\and
R. Manmatha\\
A9/Amazon\\
{\tt\small manmatha@a9.com}
\and
Alexander J. Smola\\
Amazon\\
{\tt\small smola@amazon.com}
\and
Philipp Kr\"ahenb\"uhl\\
UT Austin\\
{\tt\small philkr@cs.utexas.edu}
}
\maketitle
\thispagestyle{empty}

\begin{abstract}%
\noindent
Deep embeddings answer one simple question: How similar are two images?
Learning these embeddings is the bedrock of verification, zero-shot learning, and visual search.
The most prominent approaches optimize a deep convolutional network with a suitable loss function, such as contrastive loss or triplet loss.
While a rich line of work focuses solely on the loss functions, we
show in this paper that selecting training examples plays an equally important
role. We propose distance weighted sampling, which selects more
informative and stable examples than traditional approaches. In
addition, we show that a simple margin based loss
is sufficient to outperform all other loss functions.
We evaluate our approach on the Stanford
Online Products, CAR196, and the CUB200-2011 datasets for image retrieval and clustering, and on the LFW dataset for
face verification. 
Our method achieves state-of-the-art performance on all of them.
\end{abstract}

\section{Introduction}
\label{sec:intro}

Models that transform images into rich, semantic representations lie
at the heart of modern computer vision, with applications ranging from
zero-shot learning \cite{bucher2016improving,yuan2016hard} and visual
search
\cite{hadi2015buy,bell2015learning,song2016learnable,oh2016deep}, to
face recognition \cite{facenet, chopra2005learning,
  parkhi2015deep,sohn2016improved} or fine-grained retrieval
\cite{song2016learnable,oh2016deep,sohn2016improved}.  Deep 
networks trained to respect pairwise relationships have emerged as the
most successful embedding
models~\cite{bromley1993signature,chopra2005learning,facenet,ustinova2016learning}.

\begin{figure*}[t]
\center
\includegraphics[width=0.88\textwidth]{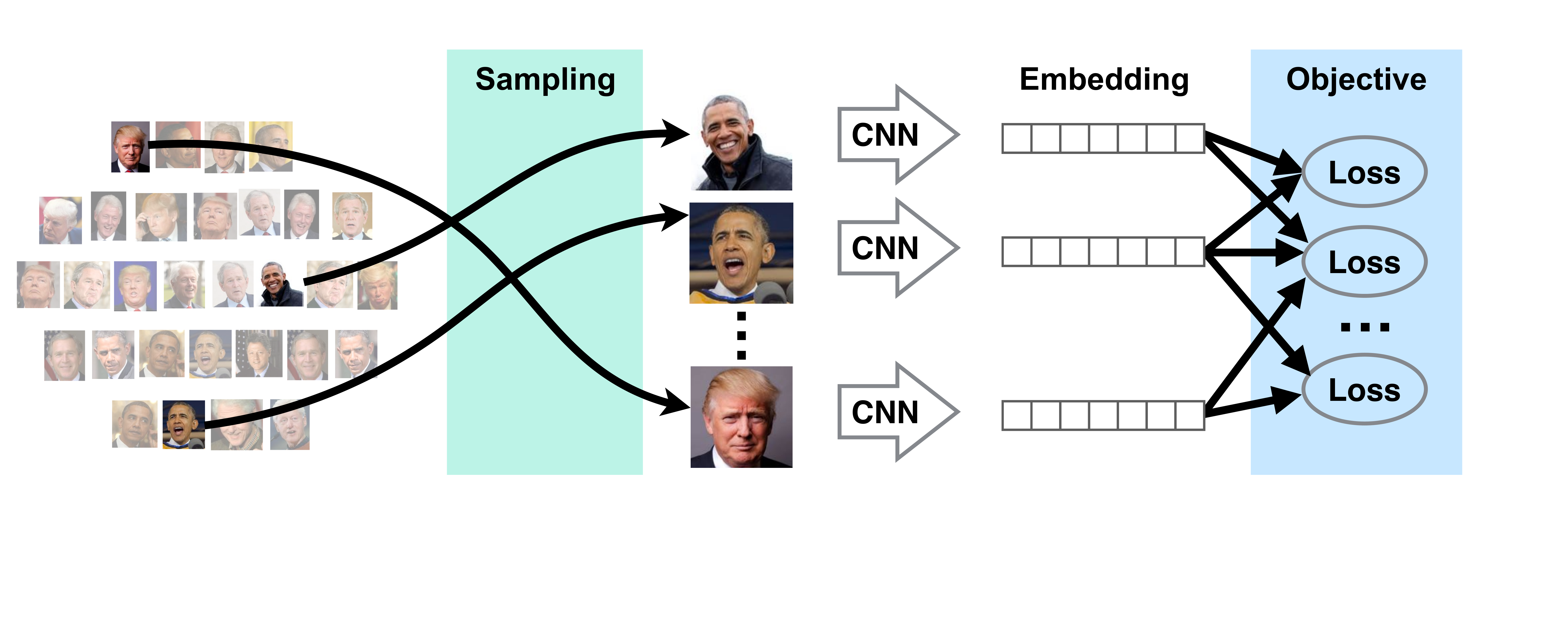}\vspace{-1mm}
\caption{An overview of deep embedding learning:
 	The first stage samples images and forms a batch. 
 	A deep network then transforms the images into embeddings.
 	Finally, a loss function measures the quality of our embedding.
 	Note that both the sampling and the loss function influence the overall training objective.
}\label{fig:teaser}
\end{figure*}

The core idea of deep embedding learning is simple:
pull similar images closer in embedding space and push dissimilar images apart.
For example, the contrastive loss \cite{hadsell2006dimensionality} forces all positives images to be close, while all negatives should be separated by a certain fixed distance.
However, using the same fixed distance for all images can be quite
restrictive, discouraging any distortions in the embedding space.
This motivated the triplet loss, which only requires negative images
to be farther away than any positive images on a per-example
basis~\cite{facenet}. 
This triplet loss is currently among the best-performing losses on standard embedding
tasks~\cite{zhuang2016fast,facenet,oh2016deep}.  
Unlike pairwise
losses, the triplet loss does not just change the loss function in
isolation, it changes the way positive and negative example are
selected. 
This provides us with two knobs to turn: the loss and the sampling strategy.
See \figref{teaser} for an illustration.



In this paper, we show that sample selection in embedding
learning plays an equal or more important role than the loss.
For example, different sampling strategies lead to
drastically different solutions for the same loss function.  At the
same time many different loss functions perform similarly under a good
sampling strategy:
A contrastive loss works almost as well as the triplet loss, if the
two use the same sampling strategy.  In this paper, we analyze
existing sampling strategies, and show why they work and why not.  We
then propose a new sampling strategy, where samples are drawn
uniformly according to their relative distance from one another.  This
corrects the bias induced by the geometry of embedding space, while at
the same time ensuring any data point has a chance of being sampled.
Our proposed sampling leads to a lower variance of
gradients, and thus stabilizes training, resulting in a qualitatively
better embedding irrespective of the loss function.

Loss functions obviously also matter.
We propose a simple margin-based loss as an extension to the
contrastive loss. 
It only encourages all positive samples to be within a
distance of each other rather than being as close as possible.
It relaxes the loss, making it more robust. 
In addition, by using isotonic regression, our margin based loss focuses on the relative orders instead of absolute distances. 


Our margin based loss and distance weighted sampling achieve
state-of-the-art image retrieval and clustering performance on the Stanford Online Products, CARS196, and the CUB200-2011 datasets.  It also outperforms previous
state-of-the-art results on the LFW face verification dataset
\cite{LFWTech} using standard publicly available training data. 
Both our loss function and sampling strategy are easy
to implement and efficient to train.

\section{Related Work}
The idea of using neural networks to extract features that respect
certain relationships dates back to the 90s.  \emph{Siamese Networks}
\cite{bromley1993signature} find an embedding space such that similar
examples have similar embeddings and vice versa. 
Such networks are trained end-to-end, sharing weights between all
mappings. Siamese Networks were first applied to
signature verification, and later extended to face verification and
dimensionality reduction \cite{chopra2005learning,
  hadsell2006dimensionality}.  However, given the limited compute
power at the time and their non-convex nature, these approaches
initially did not enjoy much attention.  Convex approaches were much
more popular~\cite{xing2002distance, comon1994independent}.  For
example, the triplet loss~\cite{schultz2003learning,
  weinberger2009distance} is one of the most prominent methods that
emerged from convex optimization.

Given sufficient data and computational power both schools of thought
were combined into a \emph{Siamese} architecture using triplet
losses. This leads to near human performance in face verification
\cite{facenet, parkhi2015deep}.  Motivated by the triplet loss, some
enforce constraints on even more examples.  For example, PDDM
\cite{huang2016local} and Histogram Loss \cite{ustinova2016learning}
use quadruplets.  Beyond that, the n-pair loss
\cite{sohn2016improved} and Lifted Structure \cite{oh2016deep} defines
constraints on all images in a batch.

This plethora of loss functions is quite reminiscent of the ranking
problem in information retrieval. There a combination of individual,
pair-wise \cite{HerGraObe99b}, and list-wise approaches \cite{CoFiRank}
are used to maximize relevance. Of note is isotonic regression which
disentangles the pairwise comparisons for greater computational
efficiency. See \cite{MoonZhengSmolaEtAl} for an overview.



Some papers explore modeling of other properties.  Structural Clustering
\cite{song2016learnable} optimizes for clustering quality.  PDDM
\cite{huang2016local} proposes a new module to model local feature
structure.  HDC \cite{yuan2016hard} trains an ensemble to model
examples of different ``hard levels". 
In contrast, here we show that a simple pairwise loss is sufficient if paired with the right sampling strategy.

Example selection techniques are relatively less studied.
For the contrastive loss it is common to select from all
posible pairs at random \cite{hadsell2006dimensionality,chopra2005learning,bell2015learning}, and sometimes with hard negative mining \cite{simo2015discriminative}. 
For the triplet loss, semi-hard negative mining, first used in FaceNet \cite{facenet}, is widely adopted \cite{oh2016deep,parkhi2015deep}. 
Sampling has been studied for stochastic optimization~\cite{zhang2015stochastic} with the goal of accelerating convergence to the same global loss function.
In contrast, in embedding learning the sampling actually changes the overall loss function considered.
In this paper we show how sampling affects the real-world performance of deep embedding learning.

\section{Preliminaries}
\label{sec:prelim}
Let $f(x_i)$ be an embedding of a datapoint $x_i \in \RR^N$, where $f: \RR^N \to \RR^D$ is a differentiable deep network with parameters $\Theta$.
Often $f(x_i)$ is normalized to have unit length for training stability \cite{facenet}.
Our goal is to learn an embedding that keeps similar data points close, while pushing dissimilar datapoints apart.
Formally we define the distance between two datapoints as 
$D_{ij} := \|f(x_i) - f(x_j)\|$, 
where $\|\cdot\|$ denotes the Euclidean norm.
For any positive pair of datapoints $y_{ij}=1$ this distance should be small, and for negative pair $y_{ij}=0$ it should be large.

The contrastive loss directly optimizes this distance by encouraging all positive distances to approach $0$, while keeping negative distances above a certain threshold:
\begin{align*}
  \contrast(i, j) := y_{ij}D_{ij}^2 + (1-y_{ij}) \sbr{\alpha - D_{ij}}_+^2.
\end{align*}
One drawback of the contrastive loss is that we have to select a
constant margin $\alpha$ for all pairs of negative samples.  This
implies that visually diverse classes are embedded in the same small
space as visually similar ones. The embedding space does not allow
for distortions.

In contrast the triplet loss merely tries to keep all positives closer to any negatives for each example:
\begin{align*}
  \triplet(a, p, n) := \sbr{D_{ap}^2 - D_{an}^2 + \alpha}_+.
\end{align*}
This formulation allows the embedding space to be arbitrarily distorted
and does not impose a constant margin $\alpha$. 

From the risk minimization perspective, one might aim at optimizing
the aggregate loss over all $O(n^2)$ pairs or $O(n^3)$ triples
respectively. That is
$$
R^{(\cdot)} := \sum_{t \in \cbr{\text{all pairs/triplets}}} \ell^{(\cdot)}(t).
$$
This is computationally infeasible.  Moreover, once the network
converges, most samples contribute in a minor way as
very few of the negative margins are violated.

This lead to the emergence of many heuristics to accelerate convergence.
For the contrastive loss, hard negative mining usually offers faster convergence.
For the triplet loss, it is less obvious, as hard negative mining often leads to collapsed models, i.e.\ all images have the same embedding.
FaceNet \cite{facenet} thus proposed to use a somewhat mysterious
\emph{semi-hard} negative mining: given an anchor $a$ and a positive
example $p$, obtain a negative instance $n$ via
$$
	n_{ap}^\star := \argmin_{n: D(a,n)>D(a,p)} D_{an},
$$
within a batch. This yields a violating example that is fairly hard but not too hard.
Batch construction also matters.
In order to obtain more informative triplets, FaceNet uses a batch
size of $1800$ and ensures that each identity has roughly $40$ images
in a batch \cite{facenet}.  
Even how to best select triplets within a batch is unclear. 
Parkhi \etal~\cite{parkhi2015deep} use online selection, so that only one triplet is sampled for every $(a,p)$ pair. 
OpenFace \cite{amos2016openface} employs offline triplet selection, so
that a batch has $\nicefrac{1}{3}$ of images as anchors, positives,
and negatives respectively. 

In short, sampling matters.
It implicitly defines a rather heuristic objective
function by weighting samples. Such an approach makes it hard to
reproduce and extend the insights to different datasets, different
optimization frameworks or different architectures. 
In the next section, we analyze some of these techniques, and explain
why they offer better results. 
We then propose a new sampling strategy that outperforms current state of the art.

\section{Distance Weighted Margin-Based Loss} \label{sec:main}



To understand what happens when sampling negative uniformly, recall
that our embeddings are typically constrained to the $n$-dimensional
unit sphere $\mathbb{S}^{n-1}$ for large $n\geq 128$. Consider
the situation where the points are uniformly
distributed on the sphere. In this case, the distribution of pairwise distances follows 
$$
  q\rbr{d} \propto d^{n-2}\sbr{1-{\textstyle \frac{1}{4}} d^2}^{\frac{n-3}{2}}.
$$
See \cite{sphere} for a derivation.
\figref{distortion} shows concentration of measure occurring.  In
fact, in high dimensional space, $q(d)$ approaches
$\Ncal(\sqrt{2},\frac{1}{2n})$.  In other words, if negative examples
are scattered uniformly, and we sample them randomly, we are likely to
obtain examples that are $\sqrt{2}$-away.  For thresholds less than
$\sqrt{2}$, this induces no loss, and thus no progress for learning.
Learned embeddings follow a very similar distribution, and thus the same reasoning applies.
See supplementary material for details.

\begin{figure}[t]
\center
\begin{tikzpicture}
\begin{axis}[
    width=0.5\textwidth,
    height=3.5cm,
    axis lines=left,
    xlabel={$D_{ij}$},
    legend pos=north west,
    legend columns=2,
    no markers,
    every axis plot/.append style={ultra thick},
    cycle multiindex* list={
      mycolorlist
          \nextlist
      mystylelist
    },
    y tick label style={
        /pgf/number format/.cd,
            fixed,
            fixed zerofill,
            precision=2,
        /tikz/.cd
    },
]
\foreach \i in {1,2,3,4,5,6}
\addplot
    table [x index=0, y index=\i,]{plotdata/distortion};
\legend{\footnotesize $n=4$,
		\footnotesize $n=8$,
		\footnotesize $n=16$,
		\footnotesize $n=32$,
		\footnotesize $n=64$,
		\footnotesize $n=128$}
\end{axis}
\end{tikzpicture}\vspace{-3mm}
\caption{Density of datapoints on the $D$-dimensional unit sphere. Note
  the concentration of measure as the dimensionality increases ---
  most points are almost equidistant.}\label{fig:distortion}
\end{figure}
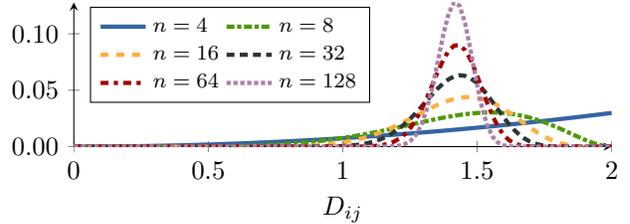

\begin{figure*}[t]
\vspace{-1.4mm}
\begin{subfigure}{0.5\textwidth}
\center
\begin{tikzpicture}
\begin{axis}[
    width=1.0\textwidth,
    height=4.8cm,
    axis lines=left,
    xlabel={Distance between examples},
    ylabel={$\nbr{\mathrm{Cov}(\nabla_{f(x_n)} \ell)}_*$},
    ymin=0.0,
    ymax=1.0,
    legend style={at={(1,1)},anchor=north east},
    legend columns=2,
    no markers,
    every axis plot/.append style={ultra thick},
    cycle multiindex* list={
      mycolorlist
          \nextlist
      mystylelist
    },
]
\foreach \i in {10,9,8,7,6,5,4,3,2,1}
\addplot
    table [x index=0, y index=\i,]{plotdata/variance};
\legend{\footnotesize $\sigma=0.1$,
		\footnotesize $\sigma=0.09$,
		\footnotesize $\sigma=0.08$,
		\footnotesize $\sigma=0.07$,
		\footnotesize $\sigma=0.06$,
		\footnotesize $\sigma=0.05$,
		\footnotesize $\sigma=0.04$,
		\footnotesize $\sigma=0.03$,
		\footnotesize $\sigma=0.02$,
		\footnotesize $\sigma=0.01$};
\end{axis}
\end{tikzpicture}\vspace{-1mm}
\subcaption{Variance of gradient at different noise levels.}\label{fig:variance}
\end{subfigure}
\hfill
\begin{subfigure}{0.5\textwidth}
\center
\begin{tikzpicture}
\begin{axis}[
    width=1.0\textwidth,
    height=4.8cm,
    ymin=0.0,
    ymax=1.0,
    axis lines=left,
    legend style={at={(1,1)},anchor=north east},
    xlabel={Distance between examples},
    every axis plot/.append style={ultra thick, mark size=4.0},
]
\addplot [aluminium4]
    table [x index=0, y index=5,forget plot]{plotdata/variance};

\addplot [maroon, only marks, mark=triangle]
	table [x index=0, y index=1]{plotdata/uniform_points};
    \addlegendentry{\footnotesize Uniform sampling}
\addplot [orange1, only marks, mark=o]
	table [x index=0, y index=1]{plotdata/hardest_points};
    \addlegendentry{\footnotesize Hard negative mining}
\addplot [chameleon3, only marks, mark=square]
	table [x index=0, y index=1]{plotdata/semihard_points};
    \addlegendentry{\footnotesize Semi-hard negative mining}
\addplot [skyblue2, only marks, mark=diamond]
	table [x index=0, y index=1]{plotdata/ours_points};
    \addlegendentry{\footnotesize Distance weighted sampling}
\end{axis}
\end{tikzpicture}\vspace{-1mm}
\subcaption{Sample distribution for different strategies.}\label{fig:sampling_methods}
\end{subfigure}\vspace{-2mm}
\caption{(a) shows the nuclear norm of a noisy gradient estimate for various levels of noise.
		 High variance means the gradient is close to random, while low variance implies a deterministic gradient estimate. Lower is better.
		 Note that higher noise levels have a lower variance at distance $0$. This is due to the spherical projection imposed by the normalization.
		 (b)~shows the empirical distribution of samples drawn for different strategies. Distance weighted sampling selects a wide range of samples, 
		 while all other approaches are biased towards certain distances.}
\end{figure*}
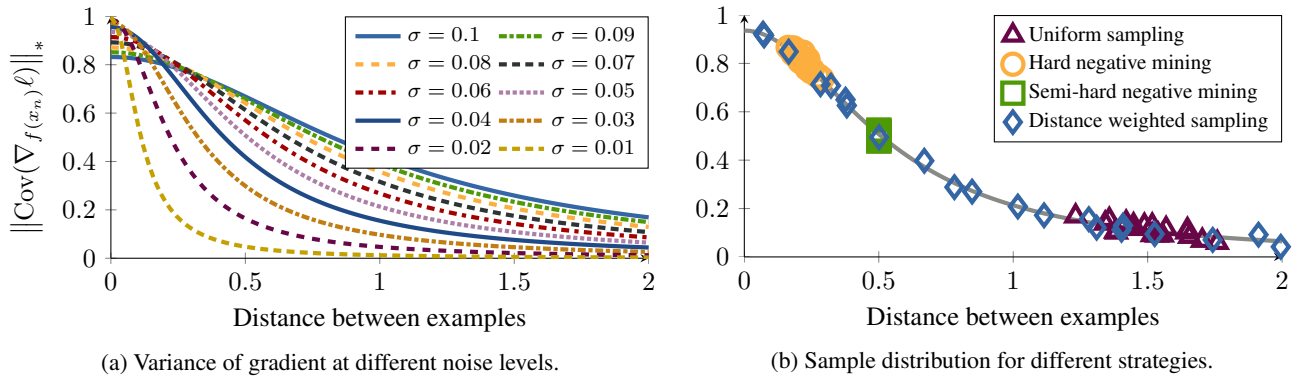

Sampling negative examples that are too hard causes a different issue.
Consider a negative pair $t:=(a, n)$ or a triplet $t:=(a, p, n)$.
The gradient with respect to the negative example $f(x_n)$ is in the form of
$$
\partial_{f(x_n)} \loss{\rbr{\cdot}} = \frac{h_{an}}{\nbr{h_{an}}}w(t)
$$
for some function $w(\cdot)$ and $h_{an} := f(x_a)-f(x_n)$.
Note that the first term $\tfrac{h_{an}}{\nbr{h_{an}}}$ determines the direction of the gradient. 
A problem arises when $\nbr{h_{an}}$ is small, and our estimates of embedding are noisy.
Given enough noise $z$ introduced by the training algorithm, 
direction $\frac{h_{an}+z}{\nbr{h_{an}+z}}$ is dominated by noise. 
\figref{variance} shows the nuclear norm of the covariance matrix for the direction of 
gradient with $z \sim \Ncal\rbr{0, \sigma^2I}$. 
We can see that when negative examples are too close/hard, the
gradient has high variance and it has low signal to noise ratio.
At the same time random samples are often too far apart to yield a good signal.

\paragraph{Distance weighted sampling.}
We thus propose a new sampling distribution that corrects the bias while controlling the variance.
Specifically, we sample uniformly according to distance, i.e.\  
sampling with weights $q(d)^{-1}$.
This gives us examples which are spread out instead of being clustered around a small region.
To avoid noisy samples, we clip the weighted sampling.
Formally, given an anchor example $a$, distance weighted sampling samples negative pair $(a, n^\star)$ with
$$
	\Pr\rbr{n^\star = n | a} \propto \min\rbr{\lambda, q^{-1}\rbr{D_{an}}}.
$$

\figref{sampling_methods} compares the simulated examples drawn from different strategies along with their variance of gradients.
Hard negative mining always offers examples in the high-variance region. 
This leads to noisy gradients that cannot effectively push two examples apart, and consequently a collapsed model.
Random sampling yields only easy examples that induce no loss. 
Semi-hard negative mining finds a narrow set in between.
While it might converge quickly at the beginning, at some point no examples are left within the band, and the network will stop making progress. 
FaceNet reports a consistent finding: the decrease of loss slows down drastically after some point, 
and their final system took 80 days to train \cite{facenet}.
Distance weighted sampling offers a wide range of examples, and thus steadily produce informative examples while controlling the variance. 
In \secref{exp}, we will see that distance weighted sampling brings performance improvements in almost all loss functions tested.
Of course sampling only solves half of the problem, but it puts
us in a position to analyze various loss functions.

\begin{figure*}[th]
\begin{subfigure}{0.25\textwidth}
	\center
	\begin{tikzpicture}[xscale=0.36,yscale=0.29]

	\draw[draw=black, line width=0.8, <->] (0,7)--(0,0)--(10,0);
	\draw[draw=black, line width=1] (7,0.2) --(7,-0.2);
	\node [below] at (7,-0.1) {$\alpha$};

	\draw[posstyle] (0,0) parabola (7,6);
	\draw[negstyle] (7,0) parabola (0,6);
	\draw[negstyle] (7,0) --(9,0);

	\end{tikzpicture}\vspace{-1mm}
	\subcaption{Contrastive loss \cite{hadsell2006dimensionality}}\label{fig:contrast}
\end{subfigure}\hfill
\begin{subfigure}{0.25\textwidth}
	\center
	\begin{tikzpicture}[xscale=0.36,yscale=0.29]

	\draw[draw=black, line width=0.8, <->] (0,7)--(0,0)--(10,0);

	\draw[draw=black, line width=1] (3,0.2) --(3,-0.2);
	\node [below] at (3,-0.1) {\small$D_{an}-\alpha$};

	\draw[draw=black, line width=1] (7.0,0.2) --(7.0,-0.2);
	\node [below] at (7.0,-0.1) {\small$D_{ap}+\alpha$};

	\draw[posstyle] (0,0)--(3,0);
	\draw[posstyle] (3,0) parabola (9,6);
	\draw[negstyle] (0,6) parabola (7,0);
	\draw[negstyle] (7,0)--(9,0);

	\end{tikzpicture}\vspace{-1mm}
	\subcaption{Triplet loss $\ell_2^2$ \cite{facenet}}\label{fig:triplet_l22}
\end{subfigure}\hfill
\begin{subfigure}{0.25\textwidth}
	\center
	\begin{tikzpicture}[xscale=0.36,yscale=0.29]

	\draw[draw=black, line width=0.8, <->] (0,7)--(0,0)--(10,0);

	\draw[draw=black, line width=1] (3,0.2) --(3,-0.2);
	\node [below] at (2.5,-0.1) {\small$D_{an}-\alpha$};

	\draw[draw=black, line width=1] (7.0,0.2) --(7.0,-0.2);
	\node [below] at (7.5,-0.1) {\small$D_{ap}+\alpha$};

	\draw[posstyle] (0,0)--(3,0);
	\draw[posstyle] (3,0)--(9,6);
	\draw[negstyle] (0,6)--(7,0);
	\draw[negstyle] (7,0)--(9,0);

	\end{tikzpicture}\vspace{-1mm}
	\subcaption{Triplet loss $\ell_2$}\label{fig:triplet_l2}
\end{subfigure}\hfill
\begin{subfigure}{0.25\textwidth}
	\center
	\begin{tikzpicture}[xscale=0.36,yscale=0.29]

	\draw[draw=black, line width=0.8, <->] (0,7)--(0,0)--(10,0);
	\draw[draw=black, line width=1] (4.5,0.2) --(4.5,-0.2);
	\draw[draw=black, line width=1] (3,0.2) --(3,-0.2);
	\draw[draw=black, line width=1] (6,0.2) --(6,-0.2);

	\draw[posstyle] (0,0) --(3,0) --(9,6);
	\draw[negstyle] (0,6) -- (6,0) --(9.3,0);

	\node [below] at (3.75,0.1) {$\alpha$};
	\node [below] at (5.25,0.1) {$\alpha$};
	\node [below] at (4.5,-0.1) {$\beta$};
	\end{tikzpicture}\vspace{-1mm}
	\subcaption{Margin based loss}\label{fig:margin_loss}
\end{subfigure}\vspace{-2mm}
\caption{Loss vs. pairwise distance. The solid blue lines show the loss function for positive pairs, the dotted green for negative pairs. Our loss finds an optimal boundary $\beta$ between positive and negative pairs, and $\alpha$ ensures that they are separated by a large margin. }
\end{figure*}
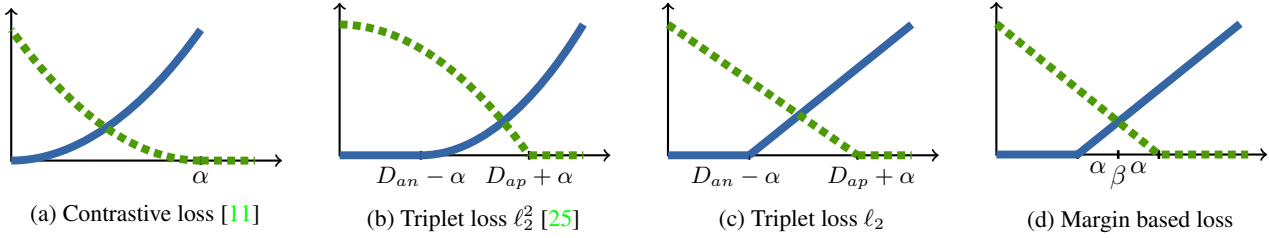

\figref{contrast} and \figref{triplet_l22} depict the contrastive loss and the triplet loss. 
There are two key differences, which in general explain
why the triplet loss outperforms contrastive loss:
The triplet loss does not
assume a predefined threshold to separate similar and dissimilar
images.  Instead, it enjoys the flexibility to distort the space to
tolerate outliers, and to adapt to different levels of intra-class
variance for different classes. 
Second, the triplet loss only requires positive examples to be closer
than negative examples, while the contrastive loss spends efforts on
gathering all positive examples as close together as possible. The latter is
not necessary.  After all, maintaining correct relative relationship
is sufficient for most applications, including image retrieval,
clustering, and verification.

On the other hand, in \figref{triplet_l22} we also observe the concave shape of the loss function for negative examples in the triplet loss.
In particular, note that for hard negatives (with small $D_{an}$), 
the gradient with respective to negative example is approaching zero. 
It is not hard to see why hard negative mining results in a collapsed model in this case: 
it gives large attracting gradients from hard positive pairs, but small repelling gradients from hard negative pairs, so all points are eventually gathered to the same point.
To make the loss stable for examples from all distances, one simple remedy is to use $\ell_2$ instead of $\ell_2^2$, i.e.
$$
	\loss{triplet,\ell_2} := \rbr{D_{ap} - D_{an} + \alpha}_+.
$$
\figref{triplet_l2} presents the loss function.
Now its gradients with respect to any embedding $f(x)$ will always have length one. 
See e.g.\ \cite{hazan2015beyond,levy2016power} for more discussions about the benefits of using gradients of a fixed length.
This simple fix together with distance weighted sampling already outperforms the traditional $\ell_2^2$ triplet loss, as shown in \secref{exp}.

\paragraph{Margin based loss.}
These observations motivate our design of a loss function which
enjoys the flexibility of the triplet loss,  has a shape suitable for
examples from all distances, while offering the
computational efficiency of a contrastive loss. The basic idea can be
traced back to the insight that in ordinal regression only the
relative order of scores matters
\cite{Joachims02}. That is, we only need to know the crossover
between both sets. Isotonic regression exploits this by estimating
such a threshold separately and then penalizes scores relative to
the threshold. We use the same trick, now applied to pairwise
distances rather than score functions. 
The adaptive margin based loss is defined as
\begin{align*}
\loss{margin}(i,j) := \rbr{\alpha + y_{ij} (D_{ij} - \beta)}_+ .
\end{align*}
Here $\beta$ is a variable that determines the boundary between
positive and negative pairs, $\alpha$ controls the margin of
separation, and $y_{ij} \in \cbr{-1, 1}$.
\figref{margin_loss} visualizes this new loss function.
We can see that it relaxes the constraint on positive
examples from contrastive loss. It effectively imposes a large margin
loss on the shifted distance $D_{ij} - \beta$.  This loss is very
similar to a support vector classifier (SVC) \cite{svm}.

To enjoy the flexibility as a triplet loss, we need a more flexible
boundary parameter $\beta$ which depends on class-specific $\bclass$ and
example-specific $\bex$ terms. 
\[\beta(i) := \bglob + \bclass_{c(i)} + \bex_i \]
In particular, the example-specific offset $\bex_i$ plays the same
role as the threshold in a triple loss. 
It is infeasible to manually select all the $\bclass_c$s and $\bex_i$s. 
Instead, we would like to jointly learn these parameters.
Fortunately, the gradient of $\beta$ can be easily calculated as
\begin{align*}
  \partial_\beta \loss{margin}(i,j) = 
  -y_{ij} \one\cbr{\alpha > y_{ij} (\beta - D_{ij})}
\end{align*}
It is clear that larger values of $\beta$ are more desirable, since
they amount to a better use of the embedding space. 
Hence, to regularize $\beta$, we incorporate a hyperparameter
$\nu$, and it leads to the optimization problem
$$
\minimize \sum_{(i,j)}\loss{margin}(i,j) + \nu \rbr{\nubglob + \nubclass_{c(i)} + \bex_i}
$$
Here $\nu$ adjusts the difference between the number of points that violate the margin on the left and on the right.
This can be seen by observing that their gradients need to cancel out at an optimal $\beta$.
Note that the use of $\nu$ here is very similar to the $\nu$-trick in $\nu$-SVM \cite{scholkopf2000new}.

\paragraph{Relationship to isotonic regression.}
Optimizing the margin based loss can be viewed as solving a ranking problem for distances.
Technically it shares similarity with learning-to-rank problems in information retrieval \cite{zheng2008query, MoonZhengSmolaEtAl}.
To see this first note at optimal $\beta$, the empirical risk can be written as 
\begin{align*}
  R^\mathrm{margin} := \min_\beta \sum_{(i,j)}\rbr{\alpha + y_{ij} (D_{ij} - \beta)}_+.
\end{align*}
One can show that $R^\mathrm{margin}=\sum_{(i,j)}\xi^*_{ij}$, where $\xi^*$s are the solution to
\begin{flalign*}
\minimize \sum_{(i,j)\in \Xcal^\mathrm{pos}} \xi_{ij} &+ \sum_{(k,l)\in \Xcal^\mathrm{neg}} \xi_{kl} \\
\text{ subject to\hspace{1.9cm}}& \\
D_{kl} + \xi_{kl} - D_{ij} + \xi_{ij} &\geq 2\alpha, 
	(i,j)\in \Xcal^\mathrm{pos}, 
	(k,l)\in \Xcal^\mathrm{neg}\\
	\xi_{ij}, \xi_{kl} &\geq 0, 
\end{flalign*}
where $\Xcal^\mathrm{pos}:=\cbr{(i,j):y_{ij}=1}$, and $\Xcal^\mathrm{neg}:=\cbr{(i,j):y_{ij}=-1}$. 
This is an isotonic regression defined on absolute error. 
We see that the margin based loss 
is the amount of ``minimum-effort" updates to maintain relative orders. 
It focuses on the relative relationships, i.e.\ focusing on the separation of positive-pair distances and the negative-pair distances.
This is in contrast to traditional loss functions such as the contrastive loss, where losses are defined relative to a predefined threshold. 


\section{Experiments}\label{sec:exp}
We evaluate our method on image retrieval, clustering and verification. 
For image retrieval and clustering, we use the Stanford Online Products \cite{oh2016deep}, CARS196 \cite{krause20133d}, and the CUB200-2011 \cite{WelinderEtal2010} datasets, following the experimental setup of Song \etal \cite{oh2016deep}.
The Stanford Online Product dataset contains 120,053 images of 22,634
categories.  The first 11,318 categories are used for training, and
the remaining are used for testing.  The CARS196 dataset contains
16,185 car images of 196 models.  We use the first 98 models for
training, and the remaining for testing.  The CUB200-2011 dataset
contains 11,788 bird images of 200 species.  The first 100 species are
used for training, the remainder for testing.  

We evaluate the quality of image retrieval based on the standard Recall@k metric, following Song \etal \cite{oh2016deep}. 
We use NMI score, $\nicefrac{I(\Omega, \CC)}{\sqrt{H(\Omega)H(\CC)}}$, to evaluate the quality of clustering alignments $\CC = \cbr{c_1, \dots, c_n}$, given a ground-truth clustering $\Omega = \cbr{\omega_1, \dots, \omega_n}$.
Here $I(\cdot, \cdot)$ and $H(\cdot)$ denotes mutual information and entropy respectively.
We use K-means algorithm for clustering.

For verification, we train our model on the largest publicly available face dataset, CASIA-WebFace \cite{yi2014learning}, 
and evaluate on the standard LFW \cite{LFWTech} dataset. 
The VGG face dataset~\cite{parkhi2015deep} is bigger, but many of its links have expired. 
The CASIA-WebFace dataset
contains 494,414 images of 10,575 people.  The LFW dataset consists of
13,233 images of 5,749 people.  Its verification benchmark contains
6,000 verification pairs, split into 10 subsets.  We select the
verification threshold for one split based on the remaining nine
splits.

Unless stated otherwise, we use an embedding size of 128 and an input image size of $224\times{224}$ in all experiments.
All models are trained using Adam \cite{kingma2014adam} with
a batch size of 200 for face verification, 80 for Stanford Online Products, and 128 for other experiments.
The network architecture follows ResNet-50 (pre-activation) \cite{he2016identity}.
To accelerate training, we use a simplified version of ResNet-50 in
the face verification experiments. 
Specifically, 
we use only $64, 96, 192, 384, 768$ filters in the 5 stages respectively, instead of the originally proposed $64, 256, 512, 1024, 2048$ filters.
We did not observe any obvious performance degradations due to
    the change. 
Horizontal mirroring and random crops from $256\times{256}$ are used for data augmentation. 
During testing we use a single center crop. 
Face images are aligned by MTCNN \cite{zhang2016joint}. 
When alignment fails, we use a center crop.
Following FaceNet\cite{facenet}, we use $\alpha=0.2$, 
and for the margin based loss we initialize $\bglob = 1.2$ and $\bclass =
\bex = 0$.

Note that some previous papers use the provided bounding boxes while others do not. 
To fairly compare with previous methods, we evaluate our methods on both the original images and the ones cropped by bounding boxes. 
For the CARS196 dataset we scale the cropped images to $256 \times 256$. 
For CUB200, we scale and pad the images such that their longer side is $256$ pixels, keeping the aspect ratio fixed.

Our batch construction follows FaceNet \cite{facenet}.
We use $m=5$ positive images per class in a batch.
All positive pairs within a batch are sampled.
For each example in a positive pair, we sample one negative pair.
This ensures that the number of positive and negative pairs are balanced, and every example belongs to the same number of positive pairs and the same number of negative pairs.

\subsection{Ablation study}

We start by understanding the effect of the loss function, the adaptive margin and
the specific functional choice.
We focus on Stanford Online Products, as
it is the largest among the three image retrieval
datasets. 
Note that image retrieval favors triplet losses over contrastive losses, since only relative relationships matter.
Here all models are trained from scratch.
Since different
methods converge at different rates, we train all methods for 100
epochs, and report the performance at their best epoch rather than at
the end of training.

We compare random sampling and semi-hard negative mining to our distance weighted sampling.
For semi-hard sampling, there is no natural choice 
of a distance lower bound for pairwise loss functions.
In this experiment we use a lower bound
of $0.5$ to simulate the positive distance in triplet loss.
We consider the contrastive loss, the triplet loss and our margin based loss.
By random sampling, we refer to uniform sampling from all positive and negative pairs. 
Since such a definition is not applicable for triplet losses, we test only the contrastive and margin based losses.

\begin{table}[t]
\centering
\begin{minipage}{\linewidth}
\small
\ra{1.0}
\setlength{\tabcolsep}{1.1em}
\begin{tabular}{@{}lcccc@{}}
	$k$&1&10&100&1000\\
	\midrule
	{\bf Random}\\
		\quad Contrastive loss \cite{hadsell2006dimensionality} &
		30.1&51.6&72.3&88.4\\
		\quad Margin
		&37.5&56.3&73.8&88.3 \\
	\cmidrule{1-5}
	{\bf Semi-hard}\\
		\quad Contrastive loss
		\cite{hadsell2006dimensionality} &
		49.4&67.4&81.8&92.1\\
		\quad Triplet $\ell_2^2$ \cite{facenet} &
		49.7&68.1&82.5&92.9  \\
		\quad Triplet $\ell_2$ & 47.4&67.5&83.1&93.6 \\
		\quad Margin &
		\underline{61.0}&\underline{74.6}&85.3&93.6 \\
	\cmidrule{1-5}
	{\bf Distance weighted} \\
		\quad Contrastive loss \cite{hadsell2006dimensionality} &
		39.2&60.8&79.1&92.2 \\
		\quad Triplet $\ell_2^2$ \cite{facenet} &53.4 &70.8&83.8&93.4 \\
		\quad Triplet $\ell_2$ &54.5&72.0 &\underline{85.4}&\bf 94.4 \\
		\quad Margin &
		\bf 61.7&\bf 75.5&\bf 86.0&\underline{94.0} \\
	\cmidrule{1-5}
		\quad Margin (pre-trained) & 72.7&  86.2&  93.8& 98.0\vspace{-2mm}
\end{tabular}
\end{minipage}
	\caption{Recall@k evaluated on Stanford Online Products.
	The bold numbers indicate the best and the underlined numbers indicate the second best performance.  
	}\label{tab:ablation}
\end{table}

Results are presented in \tabref{ablation}. 
We see that given the same loss function, different sampling
distributions lead to very different performance.  In particular,
while the contrastive loss yields considerably worse results than triplet
loss with random sampling, its performance significantly improves when
using a sampling procedure similar to triplet loss.  This 
evidence disproves a common misunderstanding of contrastive loss
vs.\ triplet loss: the strength of triplet loss comes not just from
the loss function itself, but more importantly from the accompanying sampling
methods. In addition, distance weighted sampling consistently offers
a performance boost for almost all loss functions. 
The only exception is the contrastive loss.
We found it to be very sensitive to its hyperparameters.
While we found good hyperparameters for random and semi-hard sampling, we were not able to find a well-performing hyperparameter for the distance weighted sampling yet.
On the other hand, margin based loss automatically learns a suitable offset $\beta$ and trains well. 
Notably, the margin based
loss outperforms other loss functions by a large margin irrespective of sampling strategies. 
These observations hold with multiple batch sizes, as shown in \tabref{ablation_batch}.
We also try pre-training our model using ILSVRC 2012-CLS \cite{deng2009imagenet} dataset, as is commonly done in prior work \cite{oh2016deep,bell2015learning}. 
Pre-training offers a $10\%$ boost in recall.
In the following sections we focus on pre-trained models for fair comparison.

\begin{table}[t]
\centering
\small
\ra{1.0}
\setlength{\tabcolsep}{1.1em}
\begin{tabular}{@{}llll@{}}
	Loss, batch size&Random&Semi-hard&Dist. weighted\\
	\midrule
	Triplet $\ell_2$, 40 & - &44.3&52.9\\
	Triplet $\ell_2$, 80 & - &47.4&54.5\\
	Triplet $\ell_2$, 120& - &48.8&54.7\\
	\midrule
	Margin, 40 & 41.9&60.7&\underline{61.1}\\
	Margin, 80 & 37.5&\underline{61.0}&{\bf 61.7}\\
	Margin, 120& 37.7&59.6&60.5\vspace{-2mm}
\end{tabular}
	\caption{Recall@1 evaluated on Stanford Online Products for various batch sizes (40, 80, 120). 
	Distance weighted sampling consistently outperforms other sampling strategies irrespective of the batch size.
	See supplementary material for Recall@10, 100, and 1000. 
	}\label{tab:ablation_batch}
\end{table}

\begin{figure}[t]
\small
\begin{tabular}{@{}m{0.05cm}m{1.59cm}m{2.75cm}m{2.8cm}@{}}
&Query&Triplet (R@1=49.7)&\bf Margin (R@1=61.7)
\end{tabular}\vspace{-1.3em}
\center\includegraphics[width=0.48\textwidth]{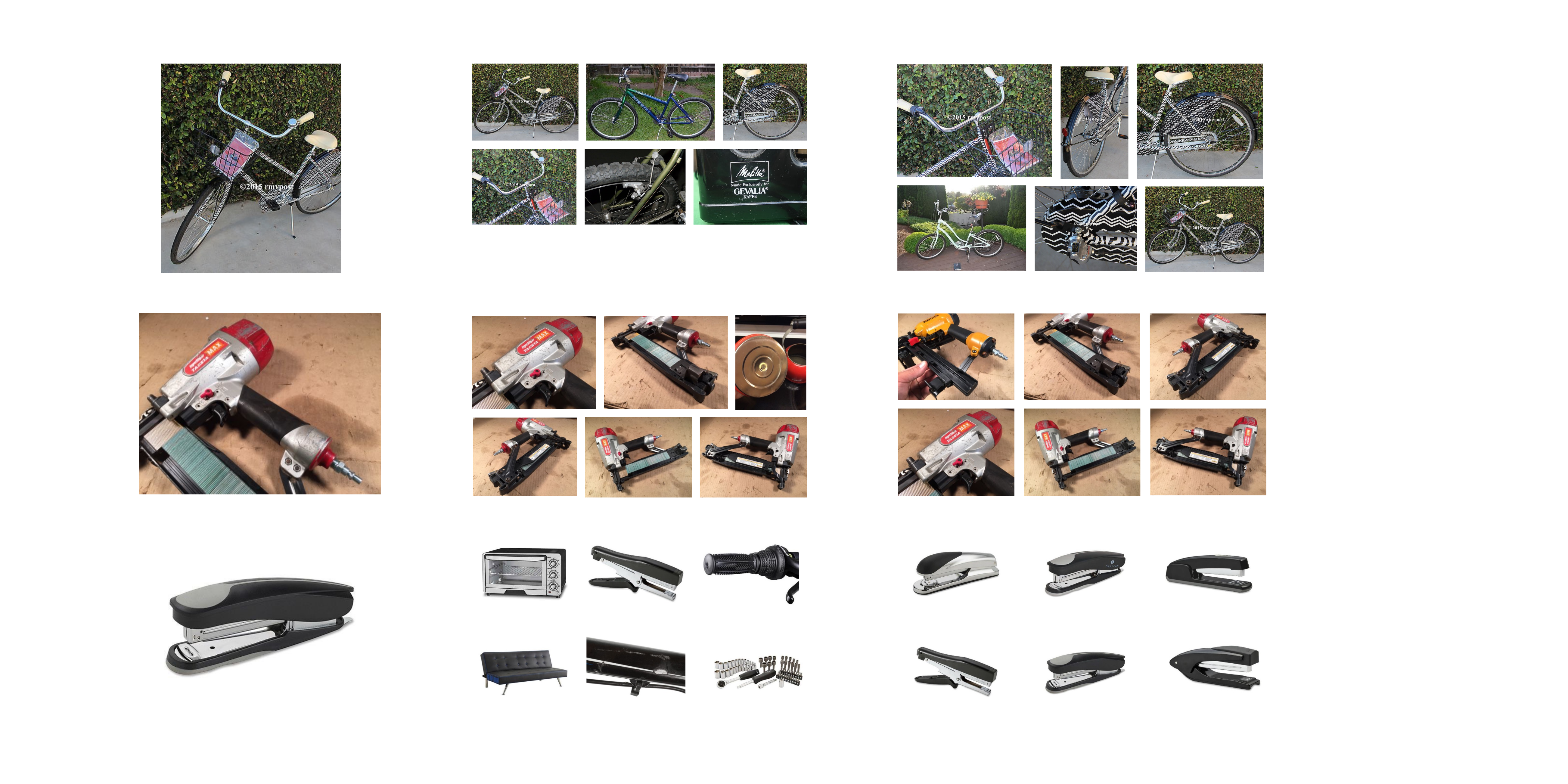}\vspace{-2.5mm}
\caption{Retrieval results for randomly chosen query images in Stanford Online Products. Our loss retrieves more relevant images. 
}\label{fig:ir}
\end{figure}

Next, we qualitatively evaluate these methods.
\figref{ir} presents the retrieval results on randomly picked query images. 
We can see that triplet loss generally offers reasonable results, but makes mistakes in some cases.
On the other hand, our method gives much more accurate results. 

To evaluate the gains obtained by learning a flexible boundary
$\beta$, we compare models using a fixed $\beta$ to models using learned $\beta$s.
The results are summarized in \tabref{beta}.  We see that
the use of more flexibly class-specific $\bclass$ indeed offers
advantages over various values of fixed $\bglob$.  We also test using
example-specific $\bex$, but the experiments are inconclusive.  We
conjecture that learning example-specific $\bex$ might have introduced
too many parameters and caused over-fitting.
\begin{table}[t]
\centering
\small
\ra{1.0}
\setlength{\tabcolsep}{1.4em}
\begin{tabular}{@{}lcccc@{}}
	$k$&1&10&100&1000\\
\midrule
	Fixed $\bglob=0.8$ &61.3&79.2&90.5&97.0\\
	Fixed $\bglob=1.0$ &70.4&84.6&93.1&97.8\\
	Fixed $\bglob=1.2$ &\underline{71.1}  &\underline{85.1}& \underline{93.2}&  \underline{97.8}\\
	Fixed $\bglob=1.4$ &67.1 & 82.6 & 92.2 & 97.7\\
	Learned $\bclass$ &\bf 72.7&\bf 86.2&\bf 93.8&\bf 98.0 \vspace{-2.5mm}
\end{tabular}\caption{Recall@k on Stanford Online Products for margin based loss with fixed and learned $\beta$. Results at 8K
  iterations are reported. 
The values of learned $\bclass_c$ range from $0.94$ to $1.45$, hence
our choice for a consensus value of $\bglob$. }
\label{tab:beta}
\end{table}

\paragraph{Convergence speed.}
\begin{figure}[t]
\center
\small 
\begin{tikzpicture}
\begin{axis}[
    width=0.5\textwidth,
    height=4.7cm,
    axis lines=left,
    xlabel={Iterations},
    xmin=0,
    xtick={0, 40, 80, 120, 160},
    xticklabels = {0, 50K, 100K, 150K, 200K},
    legend pos=south east,
    no markers,
    every axis plot/.append style={ultra thick},
    cycle multiindex* list={
      mycolorlist
          \nextlist
      mystylelist
    },
]
\addplot
    table {plotdata/accuracy_ours.dat};
    \addlegendentry{\footnotesize Margin}
\addplot
    table {plotdata/accuracy_triplet.dat};
    \addlegendentry{\footnotesize Triplet}
\addplot
    table {plotdata/accuracy_contrast.dat};
    \addlegendentry{\footnotesize Contrastive}
  
\end{axis}
\end{tikzpicture}\vspace{-2mm}
\caption{Validation accuracy curve, trained on CASIA-WebFace and evaluated on LFW. The margin based loss with distance weighted sampling converges quickly and stably, outperforming other methods.
 }\label{fig:alg_conv}
\end{figure}
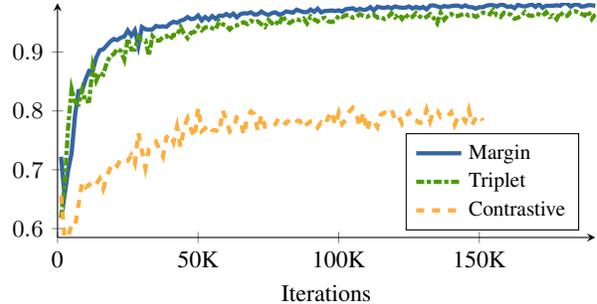

We further analyze the effects of sampling on the convergence speed.
We compare margin based loss using
distance weighted sampling with the two most commonly used deep
embedding approaches: triplet loss with semi-hard sampling and
contrastive loss with random sampling.  The learning curves are shown
in \figref{alg_conv}.  We see that triplet loss trained with semi-hard
negative mining converges slower as it ignores too many examples.  
Contrastive loss with random sampling converges even slower.  
Distance weighted sampling, 
which uses more informative and stable examples, converges faster and more accurately. 


\paragraph{Time complexity of sampling}
The computational cost of sampling is negligible. 
On a Tesla P100 GPU, forward and backward pass take about 0.55 second per batch (size 120). 
Sampling takes only 0.00031 second with semi-hard sampling and 0.0043 second with distance weighted sampling, even with our single-thread CPU implementation.
Both strategies take $\Ocal\rbr{n m (n-m)}$, where $n$ is the batch size, and $m$ is the number of images per class in a batch.

\subsection{Quantitative Results}

\begin{table}[t]
\centering
\small
\ra{1.0}
\setlength{\tabcolsep}{0.74em}
\begin{tabular}{@{}lccccc@{}}
	k&1&10&100&1000&NMI\\
\midrule
	Histogram \cite{ustinova2016learning}& 63.9 &81.7& 92.2& 97.7&-\\
	Binomial Deviance \cite{ustinova2016learning}& 65.5 &82.3 &92.3& 97.6&-\\
	Triplet Semi-hard \cite{facenet, song2016learnable}&66.7& 82.4& 91.9&-&\underline{89.5}\\
	LiftedStruct \cite{oh2016deep,song2016learnable}&62.5& 80.8& 91.9&-&88.7 \\
	StructClustering \cite{song2016learnable} &67.0& 83.7& \underline{93.2}&-&\underline{89.5}\\
	N-pairs \cite{sohn2016improved}&67.7&83.8&93.0&\underline{97.8}&88.1\\
	HDC \cite{yuan2016hard} & \underline{69.5}&\underline{84.4}&92.8&97.7&-\\
	\bf Margin &\bf 72.7&\bf  86.2&\bf  93.8&\bf 98.0&\bf90.7\vspace{-2.2mm}
\end{tabular}
\caption{Recall@k and NMI on Stanford Online
  Products~\cite{oh2016deep}. } \vspace{3mm}
\label{tab:sop}
\centering
\small
\ra{1.0}
\setlength{\tabcolsep}{0.5em}
\begin{tabular}{@{}lcccccc@{}}
	k&1&2&4&8&16&NMI\\
\midrule
&\multicolumn{6}{c}{Original Images}\\
	Triplet Semi-hard \cite{facenet, song2016learnable}& 51.5& 63.8& 73.5& 82.4&-&53.4\\
	LiftedStruct \cite{oh2016deep,song2016learnable}&53.0& 65.7& 76.0& 84.3&-&56.9\\
	StructClustering \cite{song2016learnable}& 58.1 &70.6& 80.3& 87.8&-&59.0\\
	N-pairs \cite{sohn2016improved}&71.1&79.7&86.5&91.6&-&\underline{64.0}\\
	HDC \cite{yuan2016hard} & \underline{73.7}&\underline{83.2}&\underline{89.5}&\underline{93.8}&\underline{96.7}&-\\
	\bf Margin &\bf 79.6&\bf 86.5 &\bf 91.9 & \bf 95.1 & \bf 97.3 & \bf 69.1 \\
\cmidrule{1-7}
&\multicolumn{6}{c}{Cropped Images}\\
	PDDM Triple \cite{huang2016local}  &         46.4 &58.2 &70.3 &80.1 &88.6&- \\
	PDDM Quadruplet \cite{huang2016local} &         57.4 &68.6 &80.1 &89.4 &92.3 &-\\
	HDC \cite{yuan2016hard} & \underline{83.8}&\underline{89.8}&\underline{93.6}&\underline{96.2}&\underline{97.8}&-\\
	\bf Margin
           &\bf 86.9&\bf 92.7&\bf 95.6&\bf 97.6&\bf 98.7&\bf 77.5
           \vspace{-2.2mm}
\end{tabular}
\caption{Recall@k and NMI on CARS196~\cite{krause20133d}.}  \vspace{3mm}
\label{tab:cars}
\centering
\small
\ra{1.0}
\setlength{\tabcolsep}{0.5em}
\begin{tabular}{@{}lcccccc@{}}
	k&1&2&4&8&16&NMI\\
\midrule
&\multicolumn{6}{c}{Original Images}\\
	Histogram \cite{ustinova2016learning}&52.8&64.4&74.7&83.9&90.4&-\\
	Binomial Deviance \cite{ustinova2016learning}&50.3&61.9&72.6&82.4&88.8&-\\
	Triplet \cite{facenet, song2016learnable}&42.6& 55.0& 66.4& 77.2&-&55.4\\
	LiftedStruct \cite{oh2016deep,song2016learnable}&43.6& 56.6 &68.6& 79.6&-&56.5\\
	Clustering \cite{song2016learnable}&48.2& 61.4& 71.8& 81.9&-&59.2\\
	N-pairs \cite{sohn2016improved}&51.0&63.3&74.3&83.2&-&\underline{60.4}\\
	HDC \cite{yuan2016hard} & \underline{53.6}&\underline{65.7}&\underline{77.0}&\underline{85.6}&\underline{91.5}&-\\
	\bf Margin &\bf 63.6&\bf 74.4& \bf83.1&\bf90.0&\bf94.2&\bf69.0  \\
\cmidrule{1-7}
&\multicolumn{6}{c}{Cropped Images}\\
	PDDM Triplet \cite{huang2016local} & 50.9 &62.1 &73.2 &82.5 &91.1&- \\
	PDDM Quadruplet \cite{huang2016local} & 58.3 & 69.2 &79.0 & 88.4
                   & 93.1 &- \\
    HDC \cite{yuan2016hard} & \underline{60.7}&\underline{72.4}&\underline{81.9}&\underline{89.2}&\underline{93.7}&-\\
	\bf Margin &\bf 63.9&\bf 75.3&\bf 84.4&\bf 90.6&\bf 94.8&\bf 69.8 \vspace{-2.2mm}
\end{tabular}\caption{Recall@k and NMI on CUB200-2011~\cite{WelinderEtal2010}. }
\label{tab:cub}
\end{table}

\begin{table}[h!]
\centering
\small
\setlength{\tabcolsep}{0.8em}
\ra{1.0}
\begin{tabular}{@{}lrrrr@{}}
	Model & \# training & Accuracy& Embed. & \# Nets\\
	& images & (\%) & dim.\\
\midrule
	FaceNet \cite{facenet}&200M&\bf 99.63 & 128 & 1\\
	DeepFace \cite{taigman2014deepface}&4.4M&97.35 & 4096 & 1\\
	MultiBatch \cite{tadmor2016learning}&2.6M&98.20 &128 & 1\\
	VGG \cite{parkhi2015deep} &2.6M&99.13 & 1024 & 1\\
	DeepID2~\cite{sun2014deep} & 203K& 95.43 & 160 & 1\\
	DeepID2~\cite{sun2014deep} & 203K& 99.15 & 160 & 25\\
	DeepID3~\cite{sun2015deepid3} & 300K& 99.53 &  600&25\\
\cmidrule{1-5}
	CASIA \cite{yi2014learning}&494k&97.30&320 & 1\\
	MFM \cite{wu2015lightened}&494k&\underline{98.13} &256 & 1\\
	N-pairs \cite{sohn2016improved}&494k&\underline{98.33}&320 & 1\\
	Margin&494k& \underline{98.20} &\bf  128 & 1\\
	\bf Margin&494k&\bf 98.37  &  \underline{256} & 1\vspace{-2.2mm}
\end{tabular}\caption{Face verification accuracy on LFW. We directly compare to results trained on CASIA-WebFace, shown in the lower part of the table. Methods shown in the upper part use either more or proprietary data, and are listed purely for reference. } \label{tab:lfw}
\end{table}

We now compare our approach to other state-of-the-art methods. 
Image retrieval and clustering results are summarized in \tabref{sop}, \ref{tab:cars} and \ref{tab:cub}.
We can see that our model achieves the best performance in all three datasets. 
In particular, margin based loss outperforms extensions of triplet loss, such as
LiftedStruct \cite{oh2016deep}, StructClustering~\cite{song2016learnable}, N-pairs \cite{sohn2016improved}, and PDDM \cite{huang2016local}.
It also outperforms
histogram loss \cite{ustinova2016learning}, which requires computing similarity histograms.
Also note that our model uses only one 128-dimensional embedding for each image.
This is much more concise and simpler than HDC \cite{yuan2016hard}, which uses 3 embedding vectors for each image.


\tabref{lfw} presents results for face verification. 
Our model achieves the best accuracy among all models trained on CASIA-WebFace. 
Also note that here our method outperforms models using a wide range of training procedures. 
MFM \cite{wu2015lightened} use a softmax classification loss. 
CASIA \cite{yi2014learning} use a combination of softmax loss and contrastive loss.
N-pair \cite{sohn2016improved} use a more costly loss function that is defined on all pairs in a batch. 
We also list a few other state-of-the-art results which are not comparable purely for reference.
DeepID2~\cite{sun2014deep} and DeepID3~\cite{sun2015deepid3} use 25 networks on 25 face regions based on positions of facial landmarks. 
When trained using only one network, their performance degrades significantly.
Other models such as FaceNet~\cite{facenet} and DeepFace \cite{taigman2014deepface} are trained
on huge private datasets.

Overall, our model achieves the best results on all datasets among all compared methods.
Notably, our method uses the simplest loss
function among all --- a simple variant of contrastive loss.  

\section{Conclusion}

We demonstrated that sampling matters as much or more than loss functions in deep embedding learning.
This should not come as a surprise,
since the implicitly defined loss function is (quite obviously) a
sample weighted object.  
 
Our new distance weighted sampling yields a performance improvement 
for multiple loss functions.  In addition, we analyze and provide a simple margin-based loss that relaxes
unnecessary constraints from traditional contrastive loss and enjoys
the flexibility of the triplet loss.  We show that distance weighted
sampling and the margin based loss significantly outperform all other loss
functions.
\section*{Acknowledgment}
We would like to thank Manzil Zaheer for helpful discussions.
This work was supported in part by Berkeley DeepDrive, and an equipment grant from Nvidia.

{\footnotesize
\bibliographystyle{ieee}
\bibliography{bib}
}




\clearpage
\begin{appendices}

\section{Empirical pairwise-distance distributions}
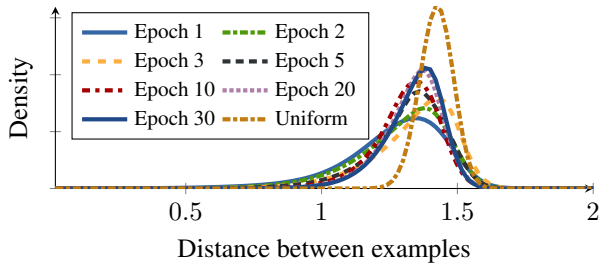
\begin{figure}[h]
\center
\begin{tikzpicture}
\begin{axis}[
    width=0.5\textwidth,
    height=4.0cm,
    axis lines=left,
    xlabel={Distance between examples},
    ylabel={Density},
    yticklabels={,,},
    legend pos=north west,
    ylabel shift = -4 pt,
    legend columns=2,
    no markers,
    every axis plot/.append style={ultra thick},
    cycle multiindex* list={
      mycolorlist
          \nextlist
      mystylelist
    },
]
\foreach \i in {1,2,3,4,5,6,7,8}
\addplot
    table [x index=0, y index=\i,]{plotdata/empirical_distribution};
\legend{\footnotesize Epoch 1,
		\footnotesize Epoch 2,
		\footnotesize Epoch 3,
		\footnotesize Epoch 5,
		\footnotesize Epoch 10,
		\footnotesize Epoch 20,
		\footnotesize Epoch 30,
		\footnotesize Uniform};
\end{axis}
\end{tikzpicture}\vspace{-2mm}
\caption{Empirical pairwise-distance distributions for negative pairs. 
They roughly follow a bell-shaped curve. }\label{fig:empirical}
\end{figure}

To better understand the effects of distance weighted sampling during training, 
we analyze our learned embeddings. 
Specifically, we compute empirical pairwise distance distributions for negative pairs based on the embeddings of testing images. 
\figref{empirical} presents the results on Stanford Online Product dataset.
We see that after the first epoch, the distribution already forms a bell shape, and in later epochs, it gradually concentrates. 
This justifies our motivation of using distance weighted sampling so that examples from all distances have a chance to be sampled.

\section{Stability analysis}
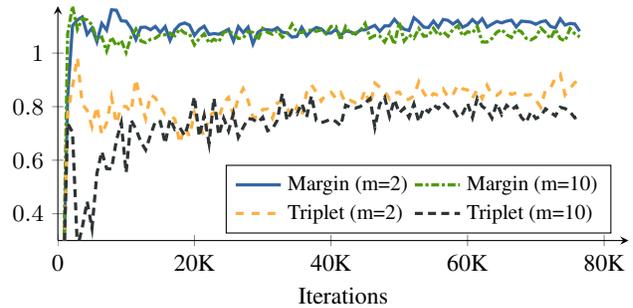
\begin{figure}[t]
\center
\small 
\begin{tikzpicture}
\begin{axis}[
    width=0.525\textwidth,
    height=4.7cm,
    xlabel={Iterations},
    legend columns=2,
    axis lines=left,
    xmin=0,
    xmax=117,
    ymin=0.3,
    xtick={0, 28, 56, 84, 112},
    xticklabels = {0, 20K, 40K, 60K, 80K},
    legend pos=south east,
    no markers,
    every axis plot/.append style={very thick},
    cycle list/Dark2-8,
    cycle multiindex* list={
      mycolorlist
          \nextlist
      mystylelist
    },
]
\addplot
    table [x index=0, y index=1] {plotdata/threshold_ours_2.dat};
    \addlegendentry{\footnotesize Margin (m=2)}
\addplot
    table [x index=0, y index=3] {plotdata/threshold_ours_2.dat};
    \addlegendentry{\footnotesize Margin (m=10)}
\addplot
    table [x index=0, y index=1] {plotdata/threshold_triplet_5.dat};
    \addlegendentry{\footnotesize Triplet (m=2)}
\addplot
    table [x index=0, y index=3] {plotdata/threshold_triplet_5.dat};
    \addlegendentry{\footnotesize Triplet (m=10)}
\end{axis}
\end{tikzpicture}\vspace{-2mm}
\caption{Optimal validation threshold for the LFW dataset. 
  Triplet loss with different sampling strategies converges to different solutions. 
  In addition, it has large fluctuations in the early stage, indicating unstable training. 
  Margin based loss always converges stably to the same solution. 
  }\label{fig:stability}
\end{figure}

Here we measure the stability of different loss functions when using
different batch construction.  Specifically, we change the number of images $m$ per class 
in a batch and see how it
impacts the solutions.  For this purpose, we experiment with face
verification and use the optimal verification boundary on the validation
set as a summary of the solution.  The results are summarized in
\figref{stability}.  We see that the triplet loss converges to
different solutions when using different batch constructions.  In
addition, we observe large fluctuations in the early stage, indicating
unstable training.  On the other hand, the margin based loss is robust, it always converges to the roughly the same geometry.

\section{Ablation study for batch size}
We analyze the sensitivity of our approach with respect to batch sizes.
\tabref{ablation_batch_all} presents the results. 
We see that distance weighted sampling consistently outperforms other sampling strategies, 
and margin based loss consistently outperforms triplet loss.
\begin{table}[t]
\centering
\small
\ra{1.0}
\setlength{\tabcolsep}{1.0em}
\begin{tabular}{@{}lcccc@{}}
    Loss, batch size&@1&@10&@100&@1000\\
    \midrule
    {Triplet $\ell_2$, 40}\\
    \quad Semihard &44.3&63.7&79.7&92.2\\
    \quad Distance weighted &52.9&70.9&83.9&94.0\\
    \midrule
    {Triplet $\ell_2$, 80}\\
    \quad Semihard &47.4& 67.5& 83.1& 93.6\\
    \quad Distance weighted &54.5& 72.0& 85.4& 94.4\\
    \midrule
    {Triplet $\ell_2$, 120}\\
    \quad Semihard &48.8&67.7&82.7&93.3\\
    \quad Distance weighted &54.7&72.7&85.9&94.6\\
    \midrule
    {Margin, 40}\\
    \quad Random &41.9&60.2&76.3&89.6\\
    \quad Semihard &60.7&75.3&85.9&94.1\\
    \quad Distance weighted &61.1&75.8&86.5&94.2\\
    \midrule
    {Margin, 80}\\
    \quad Random &37.5& 56.3& 73.8& 88.3\\
    \quad Semihard &61.0& 74.6& 85.3& 93.6\\
    \quad Distance weighted &61.7& 75.5& 86.0& 94.0\\
    \midrule
    {Margin, 120}\\
    \quad Random &37.7&56.6&73.7&88.3\\
    \quad Semihard &59.6&73.7&84.4&93.2\\
    \quad Distance weighted &60.5&74.7&85.5&93.8\\
\end{tabular}\vspace{-2mm}
    \caption{Recall@k evaluated on Stanford Online Products. 
    }\label{tab:ablation_batch_all}
\end{table}

\end{appendices}

\end{document}